\documentclass{article}
\usepackage{xcolor,soul}
\usepackage{hyperref}
\usepackage{graphicx}
\graphicspath{{./figures/}}
\usepackage{footmisc}
\usepackage{float}
\usepackage{amsthm}
\usepackage{tabularx}

\usepackage{graphicx}
\graphicspath{ {figures/} }
\usepackage{subcaption}
\usepackage{bm}

\usepackage{arxiv}

\usepackage[utf8]{inputenc} 
\usepackage[T1]{fontenc}    
\usepackage{hyperref}       
\usepackage{url}            
\usepackage{booktabs}       
\usepackage{amsfonts}       
\usepackage{nicefrac}       
\usepackage{microtype}      
\usepackage{lipsum}		
\usepackage{graphicx}
\usepackage[square,sort,comma,numbers]{natbib}
\usepackage{doi}
\usepackage{amsmath}
\usepackage{appendix}

\title{Utilising Deep Learning to Elicit Expert Uncertainty}

\author{ \href{https://orcid.org/
0000-0002-7970-6342}{Julia R. Falconer}\thanks{Julia Falconer's research is funded through the University of Waikato Doctoral Scholarship} \\
	Department of Mathematics,\\
	University of Waikato, \\
	Hamilton, New Zealand \\
	\texttt{jrg22@students.waikato.ac.nz} \\
	\And
	{Eibe Frank} \\
	Department of Computer Science,\\
	University of Waikato, \\
	Hamilton, New Zealand\\
	 \\
	\And
	{Devon L. L. Polaschek}\\
	School of Psychology,\\
	University of Waikato, \\
	Hamilton, New Zealand
	\And
	{Chaitanya Joshi}\\
	Department of Statistics,\\
	University of Auckland, \\
	Auckland, New Zealand 
}

\renewcommand{\shorttitle}{Utilising Deep Learning to Elicit Expert Uncertainty}

\hypersetup{
pdftitle={Utilising Deep Learning to Elicit Expert Uncertainty},
pdfauthor={Julia R. Falconer, Eibe Frank, Devon L. L. Polaschek, Chaitanya Joshi},
pdfkeywords={First keyword, Second keyword, More},
}

\begin{document}
\maketitle

\begin{abstract}

Recent work \cite{falconer2024eliciting} has introduced a method for prior elicitation that utilizes records of expert decisions to infer a prior distribution. While this method provides a promising approach to eliciting expert uncertainty, it has only been demonstrated using tabular data, which may not entirely represent the information used by experts to make decisions. In this paper, we demonstrate how analysts can adopt a deep learning approach to utilize the method proposed in \cite{falconer2024eliciting} with the actual information experts use. We provide an overview of deep learning models that can effectively model expert decision-making to elicit distributions that capture expert uncertainty and present an example examining the risk of colon cancer to show in detail how these models can be used. 
\end{abstract}


\pagebreak

\section{Introduction} \label{sec:intro}
	When choosing from a set of actions, it is important to understand the level of uncertainty surrounding an event in order to make informed decisions and assess risks \cite{morgan2014use}.  Analyzing numerical data or visual representations of uncertainty can assist decision-makers and risk assessors in providing insightful evaluations. However, quantifying uncertainty in this way can be a difficult task, especially where there is limited data on the event. This is where expert knowledge elicitation is needed, defined here in the statistical sense where the goal is to elicit a probability distribution that encapsulates expert uncertainty regarding some uncertain quantity or event, $E$ \cite{o2006uncertain, hartmann2020flexible}.  \\
	
	Knowledge elicitation techniques can be technical and time-consuming to apply. Standard techniques of knowledge elicitation focus on interviewing experts to obtain probability distributions reflecting their uncertainty. An analyst can interview experts and ask them questions on the probability distributions (Direct Interrogation) \cite{o2006uncertain, thomas2020probabilistic, casement2018graphical} to capture their uncertainty. Questions may be on the probabilities \cite{o2006uncertain}, such as, ``What is the probability that the proportion of $E$ is less than or equal to 0.8?"(i.e., $P(E \leq0.8)$); or focus on the quantiles of the distribution \cite{winkler1967assessment}, ``at what value is the probability of $E$ equally likely to be less than or greater than that value?" (i.e., estimate the median)\cite{falconer2021methods}. Direct Interrogation methods may induce bias from the types of questions asked (e.g. Anchoring and Adjusting, where giving a value in the question may bias the expert's response \cite{tversky1974}) and require statistical expertise on probabilities and distributions, which is often not feasible. 
	To reduce the statistical knowledge required from experts, an analyst may alternatively ask them to perform hypothetical decision-making tasks (Indirect Interrogation) \cite{eckenrode1965weighting, edwards1994smarts, wang2013expert, winkler1967quantification,falconer2021methods}. Some examples are getting experts to rank the likelihood of events or getting them to place bets on which events they think are more likely. Hypothetical decision-making tasks like these often do not place significant importance on the accuracy of the decisions, so they can potentially produce inaccuracies in the elicited distributions \cite{falconer2021methods}. \\
	
	As highlighted by \cite{o2006uncertain}, not only can expert knowledge elicitation be used for risk assessments and decision-making, it can also be profitably applied in two statistical contexts: experimental design and elicitation of informative prior distributions for Bayesian inference. In the context of Bayesian statistics, \cite{falconer2024eliciting} introduces a method for prior elicitation that combats some of the challenges faced in knowledge elicitation. Their method focuses on eliciting informative priors from expert decision-making tasks that are repetitive and carried out regularly, often under real-life circumstances. They explore the relationship between a rare event $E$ and a decision, $Y$, made by experts, selecting a decision-making task by considering whether the decision $Y$ reflects the uncertainty in the event $E$. \cite{falconer2024eliciting} illustrates this method with the event of an individual prisoner recommitting a crime upon release from prison, where there is limited data on this event occurring, and this event is hoped never to happen. A decision-making process that reflects the uncertainty around said event is the parole board decision-making process. The parole board assesses the risk of a prisoner committing a crime upon release when making their decision. Falconer et al. explain how to model the decision-making process using Bayesian inference to elicit a probability distribution capturing the uncertainty of the decision makers. Explained simply, if $Y \sim Bernoulli(p)$ and there is tabular data available from the decision-making task, then a Bayesian logistic regression model can be used to model $Y$, with $p = f(X,\beta)$, where $X$ is the information describing an event $E$. An analyst can use this model to elicit a distribution for $p$ by sampling multiple times from the posterior distribution of model parameters, $\beta$, to obtain a sample of $p$. The method of moments can be used to fit a distribution for $p$. The prior probability distribution for an event $E$ is obtained by inputting the information for the event, $X^*$, into the logistic regression model, sampling repeatedly, and using the distribution found using the method of moments as the prior probability distribution for $E$.
	This method eliminates the statistical knowledge required to elicit a prior from an expert, as the expert is just completing their usual decision making. Ideally, all decisions are made in real-life circumstances, meaning more thought is put into making the right decision. \\
	
	Although initially introduced in a Bayesian context, this method has wider applications in the field of knowledge elicitation. By relying solely on the data used for decision making instead of requiring interaction with experts, we can simplify the process of eliciting expert uncertainty. This approach is particularly useful when experts are already engaged in daily decision-making tasks. However, although promising, this method has only been investigated in the context of tabular data so far \cite{falconer2024eliciting}. Tabular data can introduce bias by only considering measurable/recorded characteristics \cite{pourhoseingholi2012control}; this means that information that is important to the decision-making process may be lost. Real-life decisions are usually made on more complex data, such as images or reports. For example, following the example in \cite{falconer2024eliciting}, the parole board considers whether a prisoner will commit a crime upon release from prison to decide whether or not to grant parole. To come up with this decision, the board reviews a report created by the prisoner's case worker and testimonials from others. This report contains all the information the board needs to make an informed decision. To obtain the appropriate distribution, we must first model the decision-making process. To do this, we must use all the information available to the decision-maker to elicit their uncertainty. Obtaining a distribution from complex data, like a report, cannot be achieved by standard statistical models. Instead, machine learning and artificial intelligence models can be utilized. In particular, deep learning is a branch of machine learning that allows analysts to model complex data. Although deep learning has been used for both uncertainty quantification \cite{hullermeier2021aleatoric} and for modelling decision-making \cite{liang2014deep}, it has never been used, as far as we are aware, to elicit expert knowledge in the form of a prior probability distribution.
	\\
	
	In this paper, we extend the work in \cite{falconer2024eliciting} using deep learning to obtain a model that captures an expert's uncertainty in the form of a probability distribution. Before delving into the specifics, it is essential to understand the fundamental concepts of uncertainty, which will be discussed in Section \ref{sec:uncertainty}. Section \ref{sec:Methods} outlines how to capture uncertainty using deep learning methods. We will then outline an example in Section \ref{sec:example} that assesses whether or not a patient is at risk of developing cancer by modelling a histopathology data set \cite{wei2021petri} with the majority diagnoses from seven pathologists. This example showcases the power of deep learning models to model complex data to obtain a probability distribution. \\

	\section{Uncertainty} \label{sec:uncertainty}
	Understanding expert uncertainty can be a challenging task. When it comes to elicitation of uncertainty, an analyst must first identify the type of uncertainty they wish to elicit and ensure that their models reflect that uncertainty. In particular, considering the method proposed in \cite{falconer2024eliciting}, it is important for the analyst to balance two factors: accurately mimicking decision-making and ensuring that the elicited distributions align with the expected behaviour with respect to the types of uncertainty deemed relevant. There are two primary types of uncertainty to consider: aleatoric uncertainty and epistemic uncertainty \cite{campos2007decision, der2009aleatory}.  Aleatoric uncertainty, also referred to as objective uncertainty, arises from random variation. An example would be tossing a fair coin. At the time of the coin toss, we can never be 100\% certain that the coin will land on heads due to the random nature of the event. Aleatoric uncertainty is irreducible, meaning it cannot be eliminated from the system. In contrast, epistemic uncertainty, also referred to as subjective uncertainty, is reducible and stems from the lack of knowledge about an event. Epistemic uncertainty occurs commonly when we have limited information on an event and is decreased when more information is provided. Although identifying what type of uncertainty is present in a particular task may seem straightforward, it is not. Identifying uncertainty is dependent on the task at hand. What is labeled as a particular type in one study may be labeled as the other type in a different study \cite{hora1996aleatory}. Kiureghian and Ditlevsen \cite{der2009aleatory} suggest that classifying uncertainty type may be more based on the immediate situation, which uncertainties can be immediately reduced and which may be more difficult to reduce in the near future. The reader should refer to \cite{der2009aleatory,hora1996aleatory} for more information.\\

	The work in \cite{falconer2024eliciting} aimed to quantify the aleatoric uncertainty of a released prisoner committing another offense by creating a model of the parole board decision-making process, which considers the uncertainty of a prisoner committing a crime after being released. To fit an appropriate model, this work explored model accuracy measures that take into account the elicited uncertainty and ensure that the model does not elicit certain distributions where it should be uncertain. The primary concern in the parole board application is to quantify aleatoric uncertainty based on the history of the prisoner. By considering parole board decisions, it may be impossible to completely eliminate epistemic uncertainty even if complete historical data on the prisoner is available because of expert bias (which is shown in the analysis performed in \cite{falconer2024eliciting}) and the difficulty of representing data comprehensively in tabular form. The goal should be to reduce the epistemic uncertainty as much as possible to highlight the aleatoric uncertainty. This may be achieved by using bias adjusting techniques or using other data types \cite{falconer2024eliciting,lichtenstein1977calibration,perala2020calibrating,russo1992managing}.\\
	
	In certain scenarios, eliciting expert epistemic uncertainty is a priority \cite{andersen2014estimating, beven2018epistemic}.  A prime example is the uncertainty surrounding a patient's diagnosis. One expert may provide a diagnosis that differs from that of another expert due to differing knowledge, experience, and training or insufficient time to form a comprehensive opinion. While obtaining the opinion of a single expert can be useful, acquiring the uncertainty of a group of experts adds crucial insight and context. If we can gather the group's uncertainty, we can better understand the uncertainty of a particular diagnosis based on the group of experts consulted. The importance of obtaining this uncertainty cannot be overstated, especially when it comes to making informed decisions about a patient's future. In Section \ref{sec:example}, an example is presented that showcases the use of eliciting epistemic uncertainty. The example involves seven experts who are diagnosing patients, but they do not always agree on a single diagnosis. We can use the expert agreement levels to check if our model is behaving appropriately in eliciting epistemic uncertainty distributions (Section \ref{sec:example}). Where we have opposing opinions, we expect our model to produce distributions that exhibit less certainty than those for patients who had a full agreement by experts. \\
	
	Our objective in this paper is to use deep learning to elicit the experts' uncertainty, taking a conservative approach when deciding whether the model is eliciting appropriate distributions. The goal should be for the model not to produce a narrow distribution where the probability mass is centred near zero or one (the model is `certain' in its outcome) in cases where the experts exhibit uncertainty about the outcome. If the opposite occurs, i.e. the model produces a distribution with a wide credible interval possibly containing 0.5 (the model is `uncertain') when all experts agree, this is less concerning and consistent with a conservative modelling approach. We would rather the model take this conservative approach than be overly certain and not follow the behaviour of the experts in cases where they are uncertain.

	In Section \ref{sec:egperformance}, we outline measures that we use to ensure our model behaves appropriately for the specified task.

	\section{Deep Learning}\label{sec:Methods}
	Understanding the human decision-making process is of interest in the field of Artificial Intelligence (AI) \cite{pomerol1997artificial}.  Deep learning is a sub-field of AI that utilizes deep neural networks to learn. A neural network (NN) structure is loosely based on the human brain and how it processes information \cite{gurney2018introduction}. It has layers of neurons that process information and pass it to the next layer until it reaches an output layer. This basic NN structure can be built upon to make more complex model architectures with many layers that can process more complex data, such as images and reports \cite{lecun1989,Goodfellow-et-al-2016}. The most common deep learning models are deterministic, but to quantify uncertainty, we need to make the deep learning models probabilistic.
	
	\subsection{Probabilistic Deep Learning}
	The key to obtaining a representation of uncertainty from a deep learning model is to apply some probabilistic features to the models. These probabilistic features are commonly used in deep learning to obtain the uncertainty surrounding model predictions by considering the uncertainty in the model parameters given the finite amount of training data available \cite{abdar2021review}. Two main approaches can be applied to deep learning models to elicit the corresponding probability distributions: Bayesian neural networks (BNN) and Monte Carlo (MC) Dropout.\\
	
	BNNs place priors on the parameters of the neural network and learn the posterior distributions of these parameters \cite{jospin2022hands}. Classical Markov Chain Monte Carlo (MCMC) methods to obtain samples from the posteriors of the parameters can be used in Bayesian deep learning, but they are computationally expensive (require large memory and a lot of time) \cite{blei2017variational} and are often challenging to implement. Applying MCMC methods to large data sets for complex networks is often impossible, given current technology.  Variational inference is more commonly used in BNNs, as it is easier to implement and does not require as much memory and time as MCMC methods \cite{blei2017variational}. Variational inference is a method to approximate the posterior distribution, $P$, by taking a distribution, $Q$, from a family of distributions of a simpler form than $P$ \cite{gelman2013bayesian}. The goal is to find a $Q$ that minimises the Kullback-Leibler divergence (Equation \ref{eq:KLD}). When applied to a neural network, variational inference finds the parameters, $\theta$ of the approximate distribution on the weights, $q(w|\theta)$ \cite{blundell2015weight}. Bayes by backdrop is a method that can be used to train a network that applies variational inference \cite{blundell2015weight}. MCMC methods and variational inference both have their specific applications in neural networks, \cite{blei2017variational} give advice on which method to use for a given deep learning task. They state: "Thus, variational inference is suited to large data sets and scenarios where we want to quickly explore many models; MCMC is suited to smaller data sets and scenarios where we happily pay a heavier computational cost for more precise samples. For example, we might use MCMC in a setting where we spent 20 years collecting a small but expensive data set, where we are confident that our model is appropriate, and where we require precise inferences. We might use variational inference when fitting a probabilistic model of text to one billion text documents and where the inferences will be used to serve search results to a large population of users". 
	\\
 \begin{equation}\label{eq:KLD}
    D_{KL}(Q||P) = \sum{}{Q log(\frac{Q}{P})}
\end{equation}
	
\begin{figure}[h]
		\centerline{
		
		\begin{subfigure}[b]{0.5\textwidth}
			\centering
			\includegraphics[width=\textwidth,height=15cm,keepaspectratio]{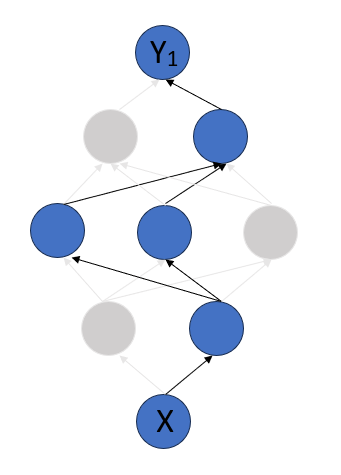}
			\caption{First pass through a NN with dropout layers.}
			\label{fig:do1}
		\end{subfigure}
		\hfill
		\begin{subfigure}[b]{0.5\textwidth}
			\centering
			\includegraphics[width=\textwidth,height=15cm,keepaspectratio]{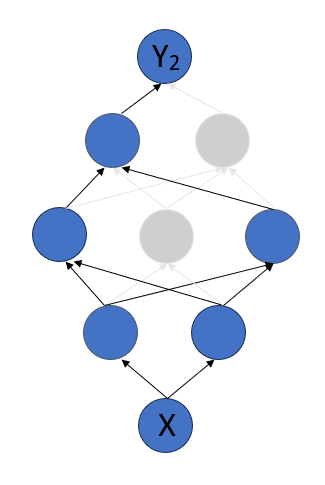}
			\caption{Second pass through a NN with dropout layers.}
			\label{fig:do2}
		\end{subfigure}}
		\caption{Multiple passes through a NN with dropout layers for a single input will produce different results.}
		\label{fig:dopass}
  
	\end{figure}

	MC-Dropout is a simple method that can produce the desired distribution. It applies random dropout to the layers of the neural network. Dropout is a function that randomly eliminates nodes from the neural network forward calculation; each forward pass will eliminate different nodes. Nodes are eliminated with a probability, $q_i$, specified as a hyperparameter by the user when the model is built. Although initially not described as a probabilistic method, MC-Dropout, in fact, provides a Bayesian approximation \cite{gal2016dropout}. Gal and Ghahramani show that dropout mathematically approximates a probabilistic Gaussian process by minimizing the Kullback-Leibler divergence between an approximate distribution and the posterior of a deep Gaussian process \cite{gal2016dropout}. An analyst can obtain a distribution by applying dropout not only at training but also when making predictions, running the input through the model multiple times, with each run randomly selecting different nodes to drop out (Figure \ref{fig:dopass}). \cite{gal2016dropout} not only show that dropout is a Bayesian approximation but, crucially, also improves model performance compared to variational inference.
	\\
	
	The key step to obtain a probability distribution from decision-making tasks through deep learning is to furnish the neural network with a mechanism to produce an output that can be viewed as a probability ( i.e., a value between zero and one). This can be achieved by applying a softmax function, $\sigma$, (Eq. \ref{Softmax}) to the outputs of the NN, $z_i$ (or a sigmoid function for binary classification). If the model outputs a probability, $p_i$, of the decision, $Y_i$, we can obtain a sample of $p_i$ through the probabilistic deep learning methods described above. Subsequently, the Method of Moments can be used, as in \cite{falconer2024eliciting}, to fit a Beta density function to the samples of $p_i$, producing a final distribution capturing uncertainty. \\
	
	\begin{equation} \label{Softmax}
		\sigma(z_i) = \frac{e ^{z_i}}{\sum^{k}_{j=1} e^{z_j}},
	\end{equation}
	
	To produce a probability distribution, analysts have various deep learning methods at their disposal. We recommend starting with a deep learning model that includes dropout layers, as it is a simpler and less computationally expensive approach that still delivers good model performance compared to other methods \cite{gal2016dropout}. It is important to note that these deep learning methods are complex models with numerous parameters; analysts must carefully consider model accuracy measures when selecting the best method while taking into account the time and memory consumed by each model. While there is still much research to be done in making these methods compatible with cutting-edge programming tools, the field of deep learning is constantly evolving, and better techniques are constantly becoming available.

	\section{Diagnosis Example}\label{sec:example}
	To make the proposed mechanism for eliciting experts' uncertainty using deep learning more concrete, we will run through a simple example with the reader to show the practical uses for prior elicitation. 
	
	\subsection{Data}
	Let $A$ be the event that an individual develops colon cancer in the future. We wish to elicit a probability distribution reflecting expert uncertainty regarding the probability of this event occurring, given imagery from histopathology. To elicit this distribution from specialists, we can model the data from the histopathology decision-making process \cite{wei2021petri}. The data set contains 3,152 images of microscope slides of colon tissue (hematoxylin and eosin (H\&E)-stained Formalin-Fixed Paraffin-Embedded (FFPE) of colorectal polyps) from the Department of Pathology and Laboratory Medicine at Dartmouth-Hitchcock Medical Center (DHMC) \cite{wei2021petri}. Each image is assigned a diagnosis of either Hyperplastic Polyp (HP) or Sessile Serrated Adenoma (SSA). The SSA diagnosis is the presence of a pre-cancerous cell which can turn into cancer if untreated \cite{makkar2012sessile,wei2021petri}. Each image was observed by seven different pathologists, and the majority vote was taken as an image's diagnosis. The data set also contains the number of pathologists who agreed with the diagnosis, showing that the diagnosis of SSA is not a certain decision and there is uncertainty among specialists in the decision. Pathologist agreement level is not included in the model building (only images and diagnosis are used) but can be used to assess model performance. There are 990 images labelled as SSA in the data set ($\approx$ 31\% of the full data set).  \\
	
	This data, and similar histopathology data sets, have been modelled by deep learning models in the past under the context of image classification \cite{wei2021petri,wang2021transpath}. The goal of these classification models is for the model to correctly identify the presence of SSA in every image. We use this data set in the context of knowledge elicitation, where the goal is to accurately quantify expert uncertainty. To do this, we can follow the basic structure of one of the models used in previous research \cite{wei2021petri} and expand this model with the probabilistic features discussed in Section \ref{sec:Methods} to obtain a distribution that captures uncertainty. After training this model, we then, in Section \ref{sec:egperformance}, complete a set of model diagnostics to make sure our model is aligned with the goal of eliciting expert uncertainty.

	\subsection{Model}
\begin{figure}[!h]
			\centering
			\includegraphics[width=\textwidth,height=20cm,keepaspectratio]{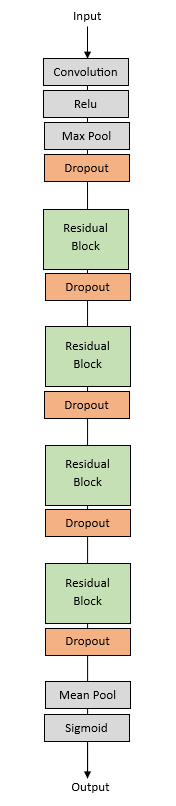}
			\caption{Basic Model Structure and Incoporation of Dropout Layers\\ }
			\label{fig:modelstructure}
		\end{figure}

	\cite{wei2021petri} models this data with a Resnet18 model. Resnet18 is a residual learning model that utilizes 18 convolutional layers, suitable for processing imagery. Residual learning solves the problem of vanishing gradients that can be found in large multi-layer networks, providing a method that will quickly train these networks and obtain better accuracy \cite{he2016deep}. Residual models contain "blocks" of small neural networks that learn features and utilise "shortcut connections" \cite{bishop1995neural}to perform identity mapping. That is, where the outputs of the previous block are added to the output of the current block\cite{he2016deep}. For more information on residual learning refer to \cite{he2016deep}. To fulfil the required probabilistic feature to output distributions, dropout layers were added between each block of the model and a sigmoid function was applied to the output layer (See Figure \ref{fig:modelstructure}). Each image was scaled down to 100 x 100, and pixel values were normalized. The initial learning rate was set to $10^{-3}$; after the first ten epochs, this was reduced by 0.01 for each epoch. The network was trained for 100 epochs. The batch size for the model was set to 32, and stochastic gradient descent was used as the optimizer.
	The model was trained with the Binary Cross Entropy (BCE) loss function in the Pytorch library \footnote{All code for model training and testing can be found at \url{https://github.com/jrg2223/Utilising-Deep-Learning-to-Elicit-Expert-Uncertainty}}.

	\pagebreak

\subsection{Model Performance}\label{sec:egperformance}
	We use the performance measures outlined in \cite{falconer2024eliciting} (Table \ref{TablePM}) to assess whether the model is appropriate for the task of uncertainty elicitation.
	To improve the reliability of the estimates of performance, the model was run ten times with different training and test sets. Accuracy readings were taken for the ten different test sets, and the average of the ten results can be found in Table \ref{tab::exaccuracy}.\\
    
	\begin{table}[!h]
		\caption{\label{TablePM} Model Diagnostics Descriptions Taken From \cite{falconer2024eliciting}}.
		\centering \small{ \centerline{
			\begin{tabular}{p{.3\textwidth}  p{.7\textwidth}}
				\hline
				\hline
				\textbf{Name}     & \textbf{Description}\\ \hline
				\hline
				\emph{Mean Accuracy} & Percentage of correct predictions the model makes by using the mean of the sampled probabilities $p_i$ for each observation.    \\
				\hline
				\emph{Mode Accuracy} & Percentage of correct predictions the model makes by using the mode of the sampled probabilities $p_i$ for each observation.\\ \hline
				\emph{Median Accuracy} & Percentage of correct predictions the model makes by using the median of the sampled probabilities $p_i$ for each observation.\\ \hline
				\emph{Area Under Curve (AUC) Accuracy} & Percentage of correct predictions the model makes by taking the largest area either side of 0.5 as the measure to form the model prediction. \\
				\hline
				\emph{95\% Credible Interval (CI) Accuracy} & Percentage of correct predictions the model makes by observing the 95\% CI of $p_i$. If the 95\% CI contains 0.5 then the assigned label can be either "Accept" or "Reject" and is a correct prediction. If the 95\% CI is contained below 0.5 and the true label is "Accept" then it is a correct prediction. If the 95\% CI is contained above 0.5 and the true label is "Reject" then it is a correct prediction.\\
				\hline
				\emph{Percentage of the 95\% CI correct predictions that contain 0.5.} & This will allow the analysts to see how many central distributions are elicited. \\
				\hline
				\emph{Percentage of the 95\% CI correct predictions that are either side of 0.5.} & This will allow the analysts to see how many skewed distributions are elicited. \\
				\hline
				\emph{F-Score \cite{sasaki2007truth}} & A measure which shows the specificity (true negative rate) and sensitivity (true positive rate) of the model. The mean of the samples of $p_i$ is used to assign labels. The highest possible value of an F-score is 1.0, indicating perfect specificity and sensitivity, and the lowest possible value is 0, if either the specificity or the sensitivity is zero. $$F = 2  \frac{specificity \times sensitivity}{ specificity + sensitivity}$$. \\
				\hline
				\emph{Confusion Matrix \cite{fawcett2006introduction}} & Shows the percentage of the mean predictions by whether the prediction is a true negative, true positive, false negative or false positive, showing the specificity and sensitivity of the model. The mean of the samples of $p_i$ is used to assign labels. \\
				\hline
				\emph{Entropy \cite{mackay2003information}} & A measure of the amount of uncertainty in a distribution. A narrow distribution will give a value close to zero, and a wide distribution will give a value closer to 1. To make sure the model is behaving correctly, it will be helpful to observe a histogram of all entropy values for the training set, as well as observe the histograms of the entropy values of correct and incorrect predictions separately. \\
				\hline
				\emph{Calibration Plot} & A calibration plot shows how well the prediction probabilities match the true percentage probabilities of the data. 
				The mean of the samples of $p_i$ is used as prediction probabilities. \\
				\hline
				\hline
		\end{tabular}}}
	\end{table}

\begin{figure}[h]
		\centerline{
		\begin{subfigure}[b]{0.6\textwidth}
			\centering
			\includegraphics[width=\textwidth,height=15cm,keepaspectratio]{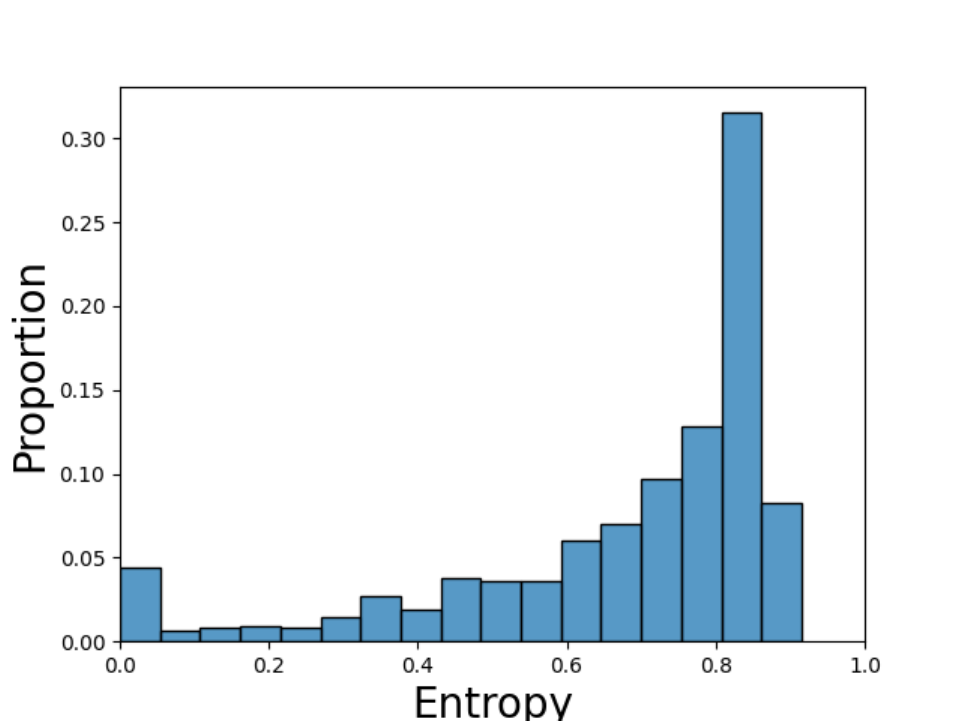}
			\caption{Histogram of the Entropy for All Test Predictions.\\ }
			\label{fig:entroeg1}
		\end{subfigure}
		}
  
		\begin{subfigure}[b]{0.5\textwidth}
			\centering
			\includegraphics[width=\textwidth,height=15cm,keepaspectratio]{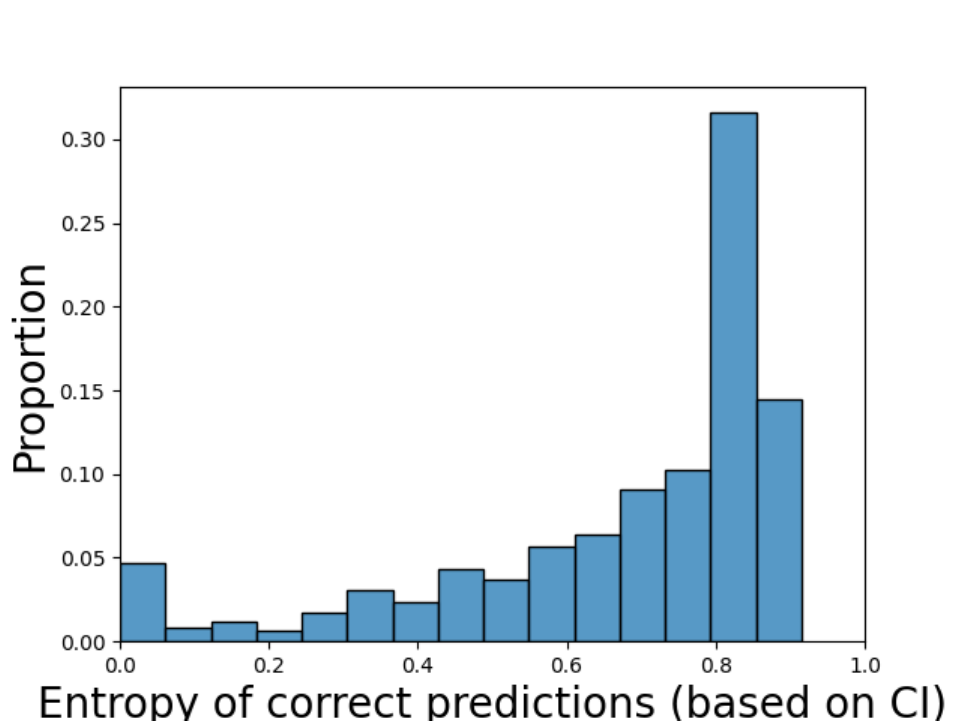}
			\caption{Histogram of the Entropy for Rest Predictions where the Model made a Correct Prediction.}
			\label{fig:corentroeg1}
		\end{subfigure}
		\hfill
		\begin{subfigure}[b]{0.5\textwidth}
			\centering
			\includegraphics[width=\textwidth,height=15cm,keepaspectratio]{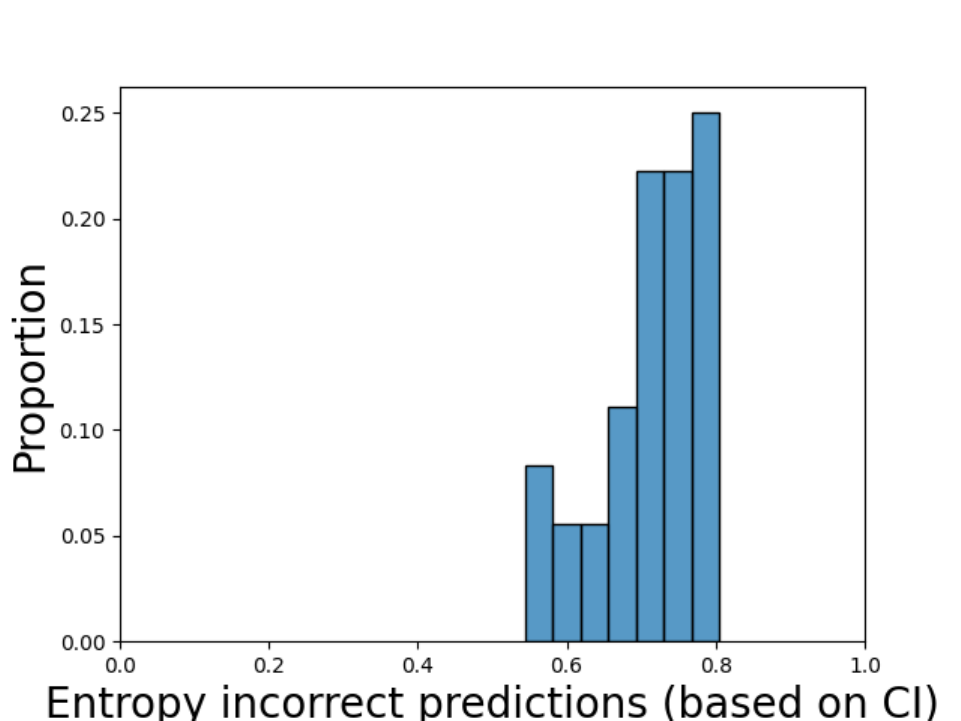}
			\caption{Histogram of the Entropy for Test Predictions where the Model made an Incorrect Prediction.}
			\label{fig:incorentroeg1}
		\end{subfigure}
		\caption{Entropy Plots for Cancer Diagnosis Example}
		\label{fig:entroplots}
	\end{figure}

\begin{figure}[!h]
 \centerline{
		\begin{subfigure}[b]{0.6\textwidth}
			
			\includegraphics[width=\textwidth,height=15cm,keepaspectratio]{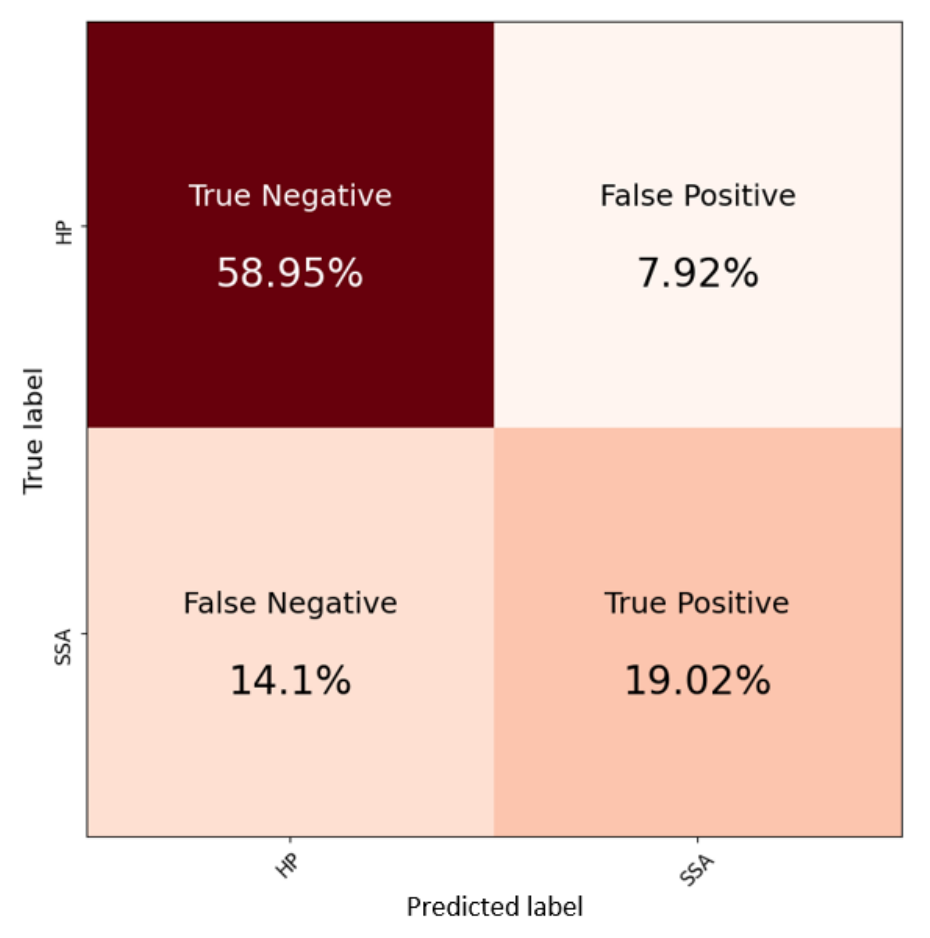}
			\caption{Confusion Matrix}\label{fig:confusion}
		\end{subfigure}
  }

  \centerline{
		\begin{subfigure}[b]{0.6\textwidth}
			\centering
			\includegraphics[width=\textwidth,height=15cm,keepaspectratio]{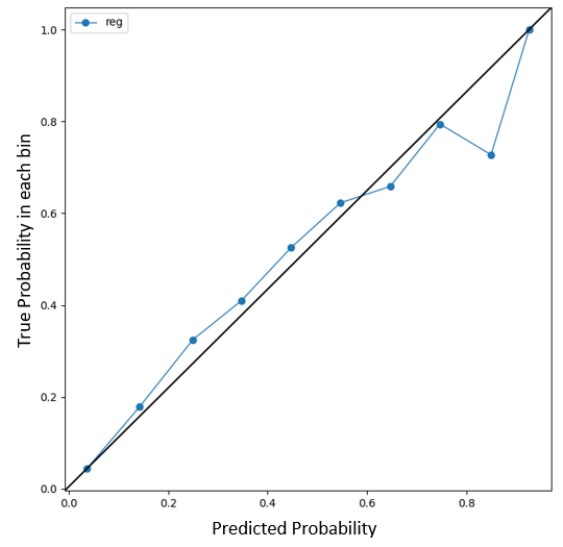}
			\caption{Calibration Plot}\label{fig:calibration}
		\end{subfigure}	}
		\caption{Model Diagnostic Plots for Cancer Diagnosis Example}\label{fig:otherplots}
	\end{figure}

	The model obtains roughly 78 \% model accuracy over mean, median, mode and AUC performance measures. For the 95 \% credible interval accuracy, we get an accuracy measure of 92.57\%; roughly 60 \% of these credible intervals are on either side of 0.5, with the other 40 \% containing 0.5. This shows us that our model is making some certain and some less certain decisions. We can also observe the behaviour of the entropy of test points in the testing data set (Figure \ref{fig:entroplots}). Entropy is a measure of uncertainty, which can give us more information on how well the model captures uncertainty. Values closer to zero indicate a narrower distribution, and values closer to one indicate a wider distribution \cite{mackay2003information}. We obtained a histogram of entropy values for all test points, a histogram of entropy values for test points that assigned the correct label by the 95\% credible interval and a histogram of entropy values that assigned the incorrect label by the 95\% credible interval. Figure \ref{fig:entroeg1} shows clearly that the model is making some certain (values close to zero) and some uncertain (values close to one) predictions. There is a peak at zero and then an exponential increase in values up to around 0.9. The histogram of correctly labeled test points shows roughly the same distribution of entropy values as the plot of all test points (Figure \ref{fig:corentroeg1}). The histogram of incorrectly labeled test points no longer has a peak at zero. Instead, a small peak at 0.4, and most points lie between 0.6 and 0.8 (Figure \ref{fig:incorentroeg1}. This shows that our model is not certain when it assigns an incorrect label, placing a wider distribution on these points. Our model is well calibrated to the data (Figure \ref{fig:calibration}) and has reasonable sensitivity and specificity (Figure \ref{fig:confusion}, and F-Score in Table \ref{tab::exaccuracy}), when using the mean of our probability samples to assign a prediction (Table \ref{tab::exaccuracy}). This model behaves reasonably well for the task of expert knowledge elicitation. \\
	
	\begin{table} [!h]
		\caption{\label{tab::exaccuracy} Average Model Performance Measures for Ten Test Data Sets.}
		\centering \centerline{
		\begin{tabular}{p{.8\textwidth} p{.15\textwidth} }
			\hline
			\textbf{Accuracy Measure}     &  \textbf{Average} \\\hline
			\hline
			\emph{Mean Accuracy} &  78.84 \%  \\
			\hline
			\emph{Mode Accuracy} & 78.59 \% \\ \hline
			\emph{Median Accuracy} & 78.91 \%  \\ 
			\hline
			\emph{AUC Accuracy} & 78.92 \%  \\
			\hline
			\emph{95\% CI Accuracy} & 92.57 \%  \\\hline
			\emph{Percentage of the 95\% CI correct predictions that contain 0.5} & 39.15 \%  \\\hline
			\emph{Percentage of the 95\% CI correct predictions that are either side of 0.5} & 60.85 \% \\\hline
			\emph{F-Score} &  0.623 \\
			\hline
		\end{tabular}}	
	\end{table}
	
	\pagebreak 
\begin{figure}[!h]
		\centerline{
		\begin{subfigure}[b]{0.6\textwidth}
			\centering
			\includegraphics[width=\textwidth,height=15cm,keepaspectratio]{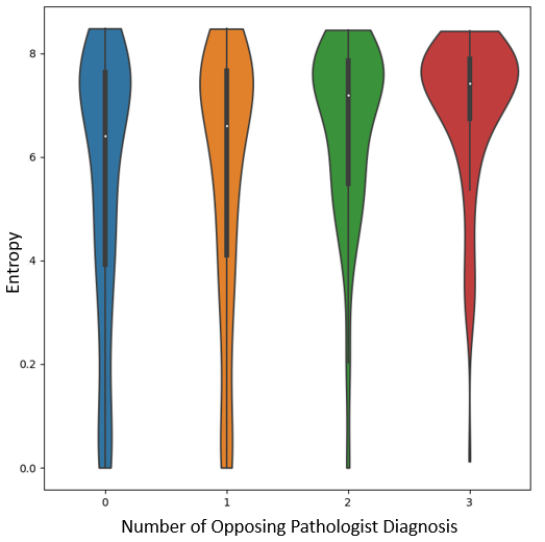}
			\caption{Distribution entropy of test dataset points by agreement level. \\  }\label{fig:agreedistent}
		\end{subfigure}	
		}
  
  \centerline{
		\begin{subfigure}[b]{0.6\textwidth}
			\centering
			\includegraphics[width=\textwidth,height=15cm,keepaspectratio]{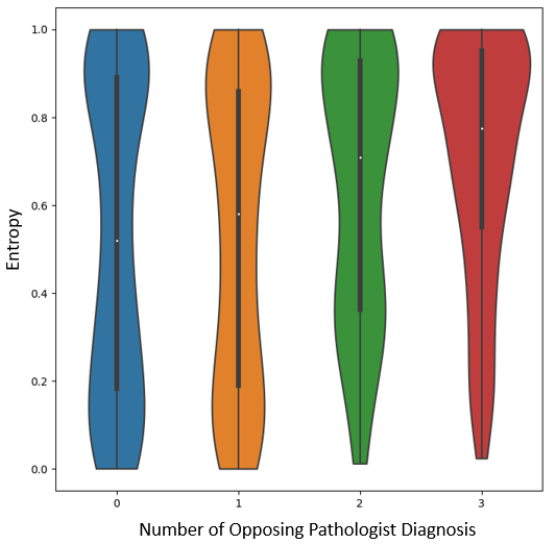}
			\caption{Mean point estimate entropy of test dataset points by agreement level. \\}\label{fig:agreemean}
		\end{subfigure}}
		\centering
		\caption{Entropy Plots of the Elicited Distributions of all Test Points. Split by Agreement Levels.}\label{fig:agreeentropyplots}
		
	\end{figure}

	\pagebreak
	We can further assess model performance by assessing the behaviour of our model based on pathologist agreement levels on the entire test set. This is often referred to as \textit{inter-rater} or \textit{inter-annotator agreement}. When modelling data that has some inter-rater disagreement, it is important to have a well-calibrated model and a model that reflects any disagreement \cite{jensen2019improving,lemay2022label}.  In our case, we have a reasonably well-calibrated model, as shown in Figure \ref{fig:calibration}. If an analyst's model is poorly calibrated, they can use methods like label smoothing to help calibrate it \cite{jensen2019improving,islam2021spatially,lemay2022label}. Additionally, we need to ensure that our uncertainty estimates reflect the varying levels of inter-rater agreement. To do this, we calculate the entropy of the probability samples for each observation in the set and also check whether its credible interval is centred. Observations are grouped into an "Agreement Level" category based on the number of pathologists who agreed on the diagnosis. If all pathologists agreed, the case is classified as "Full Agreement," if one pathologist gave a differing diagnosis, it is classified as "One Opposing," and so on. The results of this analysis are presented in Table \ref{tab::agreement}. Our findings indicate that the mean entropy increases as the number of opposing pathologists increases, indicating that overall there is greater uncertainty in the estimated distributions when the pathologist group is uncertain. Furthermore, in most cases, the number of credible intervals that are centred increases with the level of disagreement among pathologists. This is a strong indication that the model is in line with the agreement levels of pathologists, exhibiting more uncertainty in cases where the group of pathologists exhibits greater disagreement. Figure \ref{fig:agreedistent} provides further evidence of this. Violin plots are useful when displaying this type of data as they easily display the summary statistics of the data (median: white dot, interquartile range: thick black bar) while also showing the shape of the density \cite{hintze1998violin} (Figure \ref{fig:agreemean} and \ref{fig:agreedistent}). We can see that for "Full Agreement" and "One Opposing", we have thicker tails (values closer to zero) in the plot of all test set entropy values, meaning that for these classes, the model is making more certain decisions; this is reduced for "Two Opposing" and "Three Opposing" where the tail is thinner and more values are in the higher end. These results show that overall the model is exhibiting conservative behaviour, which is appropriate when eliciting uncertainty for critical decision making, as discussed in Section \ref{sec:uncertainty}. \\
	
	The entropy we have considered so far has been the entropy of the whole distribution; we shall call this \textit{distribution entropy}. Distribution entropy is important in assessing the width and height of the elicited distribution. However, there is another common entropy metric that is used for a categorical probability distribution, which can be computed by taking the mean of the sampled probabilities $p_i$ as an estimate of the probability that the (Bernoulli) event will occur, and the mean of $1-p_i$ as the estimated probability that it will not occur. If an event has a mean probability of occurring, $q$, equal to 0.5, then we can assume that the model is uncertain whether the event will occur (high entropy value). However, if $q$ is 0.9, then we can assume that it is fairly certain the event will occur  (low entropy value). We will call this \textit{point estimate entropy}. This information is also captured in stating if the credible interval contains 0.5 (Table \ref{tab::agreement}), but we can get a more specific view by looking at the point estimate entropy. We can assess the point estimate entropy for observations in each agreement level by taking the mean of the elicited distributions (Figure \ref{fig:agreemean}).  Figure \ref{fig:agreemean}  reiterates what we saw in Figure \ref{fig:agreedistent}, with the full agreement and the one opposing group having more values closer to zero than the other groups. Also, the interquartile range gets smaller, and the median of the violin plot increases as the number of opposing pathologists increases. Overall, our model is aligned with differing agreement levels.

	\begin{table}[!h] 
		\caption{\label{tab::agreement} Diagnostics for Differing Pathologist Diagnoses }
		\centering
		\begin{tabular}{l  p{.22\textwidth} p{.2\textwidth}}
			\hline
			\textbf{Agreement Level}     &  \textbf{Mean  Entropy}  & \textbf{Percentage of centred CI's}\\\hline
			\hline
			\emph{Full Agreement} & 0.6314 & 36.41 \%  \\
			\hline
			\emph{One Opposing} & 0.6433 & 32.47\% \\ \hline
			\emph{Two Opposing} & 0.7027 & 47.97\%  \\ 
			\hline
			\emph{Three Opposing} & 0.7110 & 52.08\%  \\
			\hline
			\hline
		\end{tabular}
		
	\end{table}
	
	\pagebreak
	\pagebreak
	
	\subsection{Elicited Distribution}
	After successfully training our model, we can now elicit distributions for specific individuals. This is done by inputting an image into the trained model 100 times, with dropout layers activated, to obtain a sample of 100  probabilities. The Method of Moments \cite{pearson1936method} is used to fit a beta distribution to the samples.
	\\
	
	To show examples of elicited distributions, we chose four individuals from the data set and excluded them from any of the training and test data sets. Two of them were diagnosed with SSA, while the other two had HP. Images from each of these individuals were passed through the trained model, and distributions were obtained. We further evaluated our model's performance by selecting two individuals for each diagnosis based on the pathologists' agreement. One individual had a high number of pathologists agreeing on the diagnosis (seven out of seven), while the other had a low agreement (four out of seven). For individuals with high agreement, we expect the elicited distributions to resemble a certain expert decision, with a narrower distribution on either side of 0.5. Conversely, for individuals with low agreement, we expected our distributions to be wider and more centred around 0.5. \\
	
\begin{figure}[h]
 \centerline{
		\begin{subfigure}[b]{0.5\textwidth}
			\centering
			\includegraphics[width=\textwidth,height=15cm,keepaspectratio]{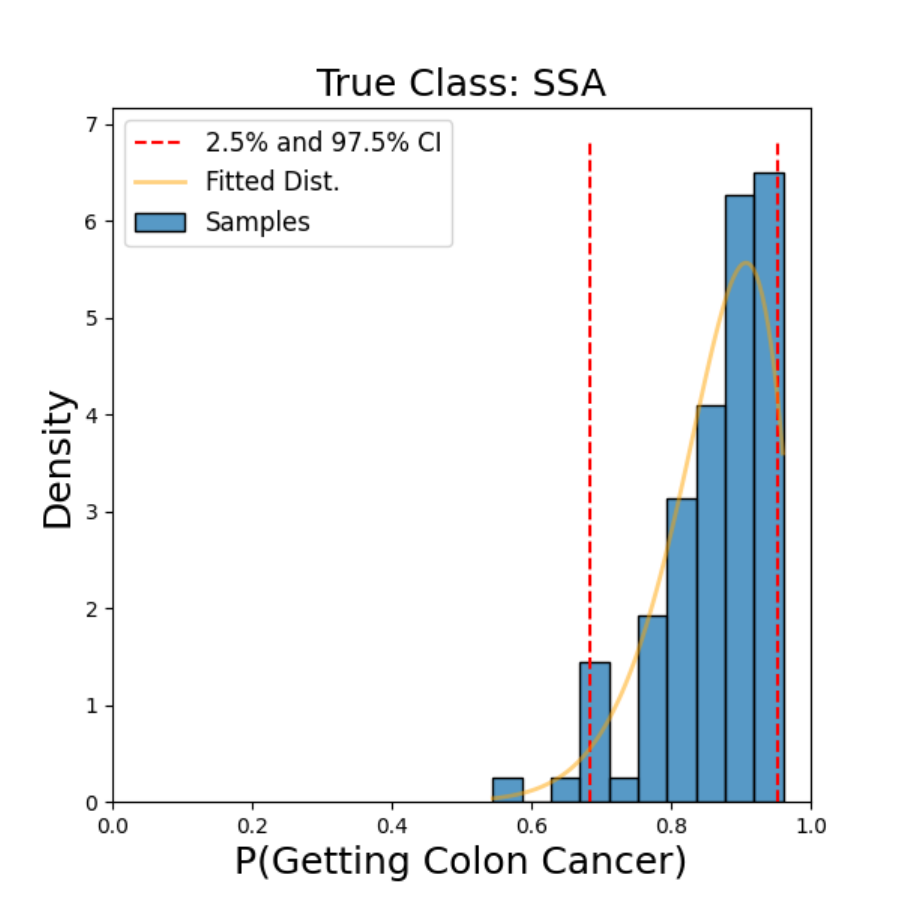}
			\caption{Individual A: diagnosed with SSA where seven out of seven pathologists diagnosed SSA}
			\label{fig:ssahighagreeA}
		\end{subfigure}
		\begin{subfigure}[b]{0.5\textwidth}
			\centering
			\includegraphics[width=\textwidth,height=15cm,keepaspectratio]{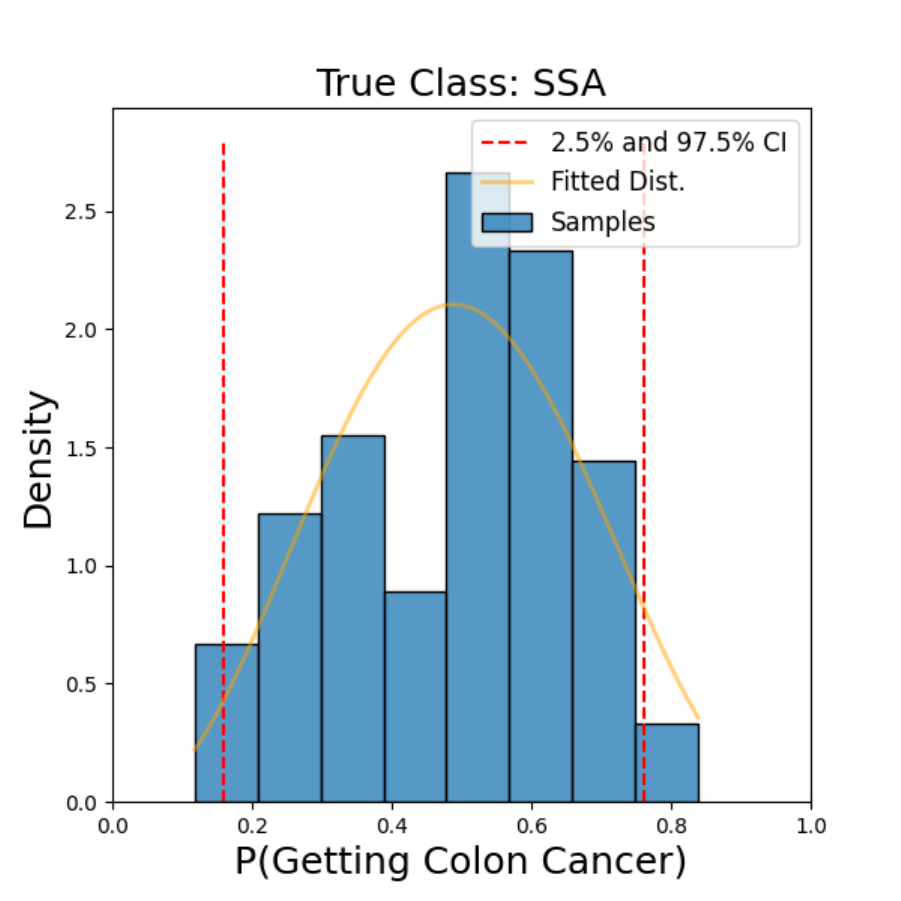}
			\caption{Individual B: diagnosed with SSA where four out of seven pathologists diagnosed SSA}
			\label{fig:ssalowagreeB}
		\end{subfigure}}
		\caption{Elicited Prior Probability Distributions for Individuals Diagnosed with SSA}
		\label{fig:ssaplot}
	\end{figure}

\begin{figure}[h]
 \centerline{
		\begin{subfigure}[b]{0.5\textwidth}
			\centering
			\includegraphics[width=\textwidth,height=15cm,keepaspectratio]{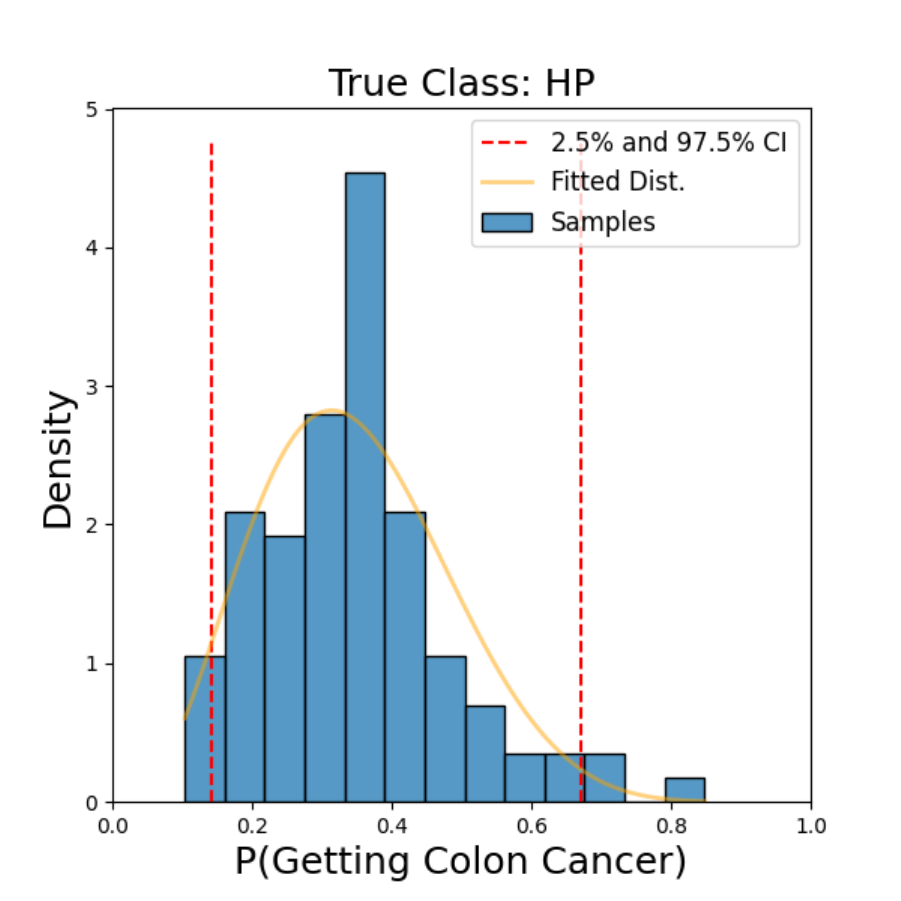}
			\caption{Individual C: diagnosed with HP where seven out of seven pathologists \\ diagnosed HP}
			\label{fig:hphighagreeC}
		\end{subfigure}
		\begin{subfigure}[b]{0.5\textwidth}
			\centering
			\includegraphics[width=\textwidth,height=15cm,keepaspectratio]{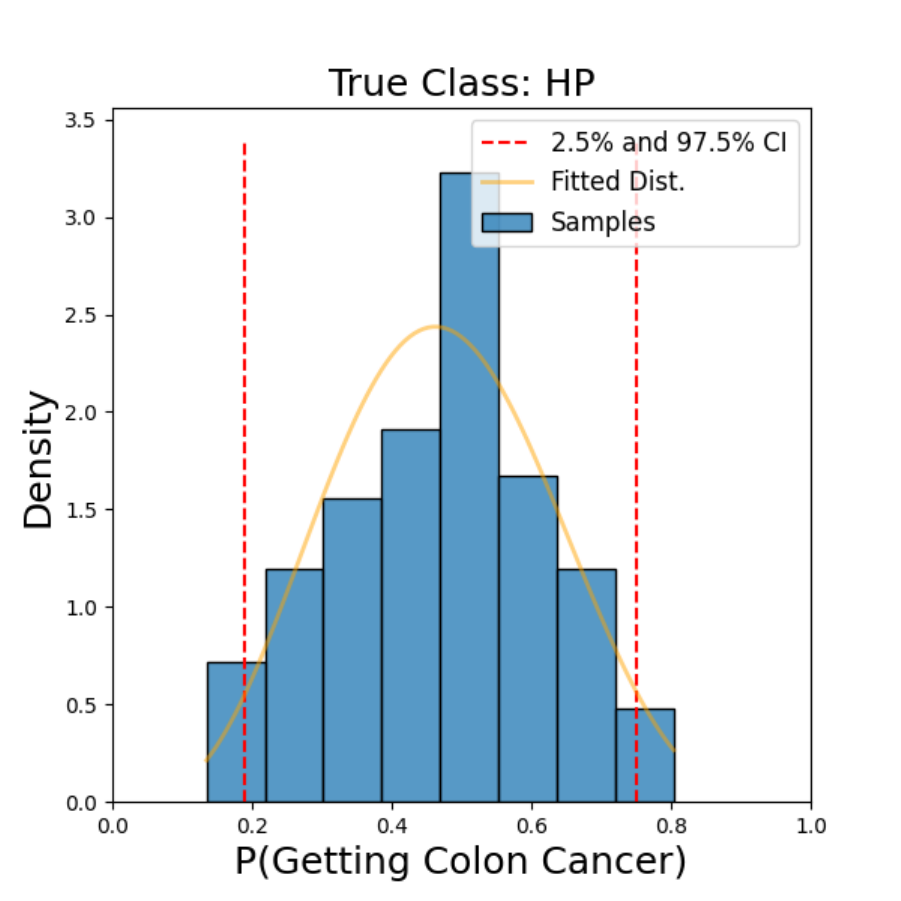}
			\caption{Individual D: diagnosed with HP where four out of seven pathologists \\ diagnosed HP}
			\label{fig:hplowagreeD}
		\end{subfigure}}
  \centering
		\caption{Elicited Prior Probability Distributions for Individuals Diagnosed with HP}
		\label{fig:hpplot}
	\end{figure}

	For Individual A, who was given the same diagnosis by all seven pathologists, our model elicited a $Beta(15.718,2.502)$, and the probability distribution was left-skewed and relatively narrow (Figure \ref{fig:ssahighagreeA}). In contrast, Individual B, given the same diagnosis by only four out of seven pathologists, had a more centred and wider distribution (fitted $Beta(3.656,3.784)$). These results show that our model's predictions for SSA diagnosis align with pathologists' uncertainty in diagnoses. For patients given the HP diagnosis, there are similar results, albeit perhaps exhibiting slightly more uncertainty than for individuals with the SSA diagnosis. The distribution elicited for Individual C (fitted $Beta(3.953,7.450)$) is less central than the elicited distribution for Individual D (fitted $Beta(4.591,5.174)$), with the Individual C distribution being slightly left skewed. Both distributions have similar widths (Figure \ref{fig:hpplot}).
	
	\section{Conclusions and Future Research}\label{sec:conclusion}
	This article demonstrates the usage of the method for prior elicitation presented in \cite{falconer2024eliciting} in the wider context of eliciting expert uncertainty, particularly for decision-making tasks with complex data like reports and images. We illustrate how by introducing probabilistic features in deep learning models, we can use records of expert decision-making to elicit expert uncertainty. In particular, applying dropout layers to a deep learning model is a simple yet effective method to obtain a probability distribution that encapsulates experts' uncertainty.  As an example, we use dropout to elicit a distribution for estimating the risk of cancer in a patient's future. This example showcases how deep learning models can imitate decision making and capture experts' uncertainty. The dropout model performs well by eliciting both certain and uncertain distributions that align with the experts' uncertainty. The goal is for these models to conservatively capture expert uncertainty in the elicited distributions.\\
	
	Our example highlights the simplicity of using historical decision making data for uncertainty elicitation in healthcare. Medical literature typically uses expert elicitation methods that involve interviewing experts about distribution probabilities and quantiles \cite{azzolina2021prior,johnson2010valid}. Analysts often set up the elicitation process using established protocols such as the SHELF protocol \cite{gosling2018shelf,azzolina2021prior,graziadio2020uncertainty}. These methods tend to take a lot of time to produce sufficient results, as a considerable amount of preparatory work must be undertaken before the interviews with experts can proceed \cite{graziadio2020uncertainty,rossi2019expert}. Getting accurate results from interviews takes time and caution. First, an expert must be trained in probability theory, and then time must be set aside for experts to be interviewed and followed up with if required \cite{graziadio2020uncertainty}. Then analysts must take precautions when asking questions, to ensure the elicitation process does not induce bias \cite{johnson2010valid}. Modelling decision making data to elicit distributions requires little preparatory work, only for data to be collected and models to be built. It also does not induce bias through the elicitation process itself. This provides a simpler option for implementation compared to typical methods. It is worth noting when considering applications to healthcare, the field of knowledge elicitation would benefit from more research into comparing currently available techniques. It would be useful to see the exact differences (if any) in distributions elicited through other methods and those elicited from expert decision making.\\
	
	Expert knowledge elicitation is often a key component in critical and strategic decision-making. As such, our proposed deep learning based expert elicitation approach can be used in decision-making approaches based on decision theory, risk analysis or adversarial risk analysis (ARA) \cite{rios2009adversarial}. This is especially so where the decision-making is  repetitive and frequent, yet each case is unique in its own way and the decision has important implications. ARA, in particular, is based on game Theory principles and aims to find the optimal actions for a defender, while taking into account prior uncertainty surrounding an offender; any uncertainty must be elicited from experts before ARA can be applied. The method presented in this paper can be used to elicit these distribution for ARA models for repetitive decision-making as highlighted in \cite{joshi2024parole}. Yet, the potential applications are beyond those in the justice and medical diagnosis fields and also includes applications related to actuarial and security risks, for example. Future research should be undertaken to show how the proposed expert elicitation approach can be used in such models.\\
	
	Lastly, it is important to note that deep learning is a proliferating field with many different research avenues. Analysts should expect research to significantly improve models in the near future, which will make applying deep learning models to knowledge elicitation tasks easier. 

\bibliographystyle{plain}
\bibliography{Main}

\end{document}